\documentclass{article}

\usepackage[preprint]{neurips_2025}

\usepackage[utf8]{inputenc} 
\usepackage[T1]{fontenc}    
\usepackage{hyperref}       
\usepackage{url}            
\usepackage{booktabs}       
\usepackage{amsfonts}       
\usepackage{nicefrac}       
\usepackage{microtype}      
\usepackage{xcolor}         
\usepackage{algorithm}
\usepackage{algorithmic}
\usepackage{bm}
\usepackage{multirow}
\usepackage{multicol}
\usepackage{enumitem}
\usepackage{colortbl}
\usepackage{tabularx}
\usepackage{makecell}
\usepackage{algorithm}
\usepackage{algorithmic}

\usepackage{pifont}
\usepackage{graphicx}
\usepackage{amsmath,amssymb}
\usepackage{amsthm}
\usepackage{hhline}
\usepackage{float}
\usepackage{array}
\newtheorem{remark}{Remark}
\newtheorem{theorem}{Theorem}[section]
\definecolor{tableheadcolor}{RGB}{255, 255, 255}
\definecolor{tablerowcolor}{RGB}{242, 242, 242}
\definecolor{ourrowname}{RGB}{255, 255, 255}

\definecolor{mydarkblue}{rgb}{0,0.08,0.45}

\definecolor{mydarkblue}{rgb}{0,0.4,0.8} 
\definecolor{mylightblue}{rgb}{0.5,0.75,1}
\definecolor{NavyBlue}{HTML}{000080}
\hypersetup{
    colorlinks=true,
    linkcolor=red,
    citecolor=NavyBlue,
    filecolor=magenta,      
    urlcolor=magenta,
}
\def\method{DENSE}
\def\methodlower{dynamic text bundling supervision}

\title{Dynamic Bundling with Large Language Models for Zero-Shot Inference on Text-Attributed Graphs}

\author{%
  Yusheng Zhao\textsuperscript{1}, Qixin Zhang\textsuperscript{2}, Xiao Luo\textsuperscript{4}, Weizhi Zhang\textsuperscript{5}, \\ \textbf{Zhiping Xiao\textsuperscript{3}, Wei Ju\textsuperscript{1}, Philip S. Yu\textsuperscript{5}, Ming Zhang\textsuperscript{1}} \\
  \textsuperscript{1} Peking University, \textsuperscript{2} Nanyang Technological University, \textsuperscript{3} University of Washington \\ \textsuperscript{4} University of California, Los Angeles,
  \textsuperscript{5} University of Illinois Chicago, \\
\texttt{yusheng.zhao@stu.pku.edu.cn}, \texttt{qixinzhang1106@gmail.com} \\
\texttt{xiaoluo@cs.ucla.edu},
\texttt{\{wzhan42, psyu\}@uic.edu},\\
\texttt{patxiao@uw.edu}, 
\texttt{\{juwei,mzhang\_cs\}@pku.edu.cn} \\
}

\begin{document}

\maketitle

\begin{abstract}
Large language models (LLMs) have been used in many zero-shot learning problems, with their strong generalization ability. Recently, adopting LLMs in text-attributed graphs (TAGs) has drawn increasing attention. However, the adoption of LLMs faces two major challenges: \emph{limited information on graph structure} and \emph{unreliable responses}. LLMs struggle with text attributes isolated from the graph topology. Worse still, they yield unreliable predictions due to both information insufficiency and the inherent weakness of LLMs (\emph{e.g.}, hallucination). Towards this end, this paper proposes a novel method named \underline{D}ynamic T\underline{e}xt Bu\underline{n}dling \underline{S}up\underline{e}rvision (\method{}) that queries LLMs with bundles of texts to obtain bundle-level labels and uses these labels to supervise graph neural networks. Specifically, we sample a set of bundles, each containing a set of nodes with corresponding texts of close proximity. We then query LLMs with the bundled texts to obtain the label of each bundle. Subsequently, the bundle labels are used to supervise the optimization of graph neural networks, and the bundles are further refined to exclude noisy items. To justify our design, we also provide theoretical analysis of the proposed method. Extensive experiments across ten datasets validate the effectiveness of the proposed method.
\end{abstract}

\section{Introduction}
\label{sec:intro}
Text-attributed graphs (TAGs) \cite{yan2023comprehensive, zhang2024text} are an important form of graph data, containing textual descriptions associated with each node. By combining textual information with non-Euclidean graph topology, TAGs serve as natural structured data representation in many applications, including citation networks \cite{radicchi2011citation}, social networks \cite{nettleton2013data}, e-commerce networks \cite{liu2021item}, and webpage networks \cite{craven1998learning}. 
As complete labeling of these large networks is often time-consuming and costly, efforts have been made to utilize semi-supervised learning \cite{kipf2016semi, verma2021graphmix}, transfer learning \cite{dai2022graph, zhu2025graphclip}, and few-shot/zero-shot learning \cite{ding2020graph, zhao2024pre, yu2025leveraging} to understand text-attributed graphs with limited labels.

Large language models (LLMs) \cite{yang2024qwen2, liu2024deepseek} have been observed to exhibit strong zero-shot generalization capability, enhancing the performance on various types of data, including visual signals \cite{zhao2025enhancing}, texts \cite{liu2024liberating}, programming code \cite{ye2025uncovering}, and graphs \cite{wang2024llms}. 
Recently, there have been efforts in integrating LLMs in text-attributed graphs \cite{he2023harnessing, huang2024gnns, zhu2025llm}. 
One line of research integrates the graph topology into language models \cite{chen2024text, mao2024position, zhu2025llm}, converting non-Euclidean topology into a sequence of tokens. However, building such foundation models requires a large amount of data \cite{wang2025model}, and the conversion to Euclidean data inevitably incurs information loss \cite{liu2025graph}.
Another line of research directly utilizes the zero-shot generalization ability of existing LLMs to understand node attributes \cite{chen2024exploring, li2024similarity, wang2025model}, and utilizes the output of LLMs as supervision signals for training graph neural networks (GNNs) \cite{kipf2016semi} or as clustering centers \cite{wang2025model}. However, the text attributes are often isolated from the graph topology, and the unreliable responses from LLMs also pose challenges for subsequent operations.

\begin{figure}
    \centering
    \includegraphics[width=0.95\linewidth]{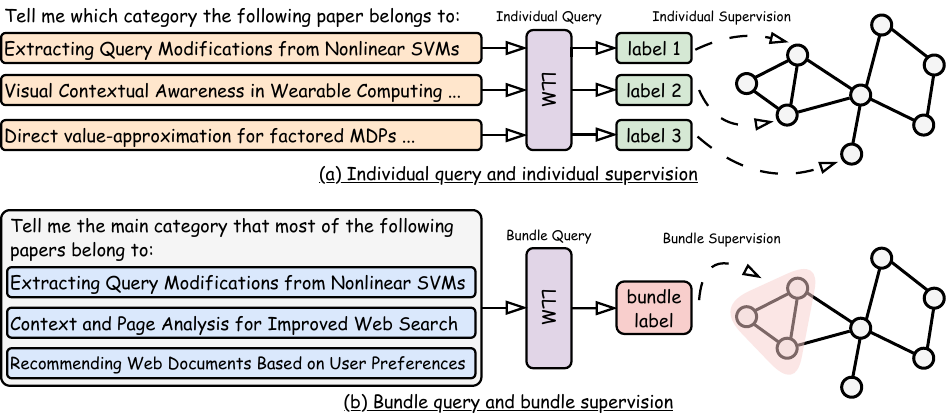}
    \vspace{-2mm}
    \caption{(a) Querying LLMs with individual texts and supervising graph learning with individual labels. (b) By creating text bundles, we perform bundle queries to obtain bundle labels for supervision.}
    \vspace{-5mm}
    \label{fig:bundle}
\end{figure}

Adopting LLMs in zero-shot inference on TAGs faces two major challenges: \emph{(1) LLMs receive limited information on graph structure.} Graph topology is non-Euclidean, making it difficult to transform into token sequences with limited context windows. \emph{(2) LLMs yield unreliable responses.} The inherent weakness of LLMs (\emph{e.g.}, hallucination), together with limited information, makes the responses from LLMs unreliable, damaging subsequent operations like clustering, classification, or supervision.

Towards this end, this paper proposes a novel method named \methodlower{} that queries LLMs and supervises graph neural networks using text bundles. As can be seen in Figure \ref{fig:bundle}, conventional methods \cite{chen2023label, wang2025model} query LLMs with individual text items (\emph{e.g.}, in citation networks, this would be individual papers' titles and abstracts). The LLMs then return the annotations of these texts, which are used as supervision signals. This paradigm faces the two major challenges mentioned above: the LLMs suffer from limited information, and the downstream supervision signals are unreliable. By comparison, this work proposes to query LLMs and supervise subsequent graph learning with text bundles. We first sample topologically or semantically similar text items to form a text bundle, and then query the LLMs about the \emph{mode category} (\emph{i.e.}, the most frequent category of the text items) as the bundle label. Subsequently, we design bundle supervision that uses the bundle labels to train a graph neural network, and during this process, bundles are further refined to exclude noisy items. In this way, the LLMs receive richer information from multiple interrelated text items in a bundle (challenge 1), and the predicted bundle labels are more robust to the uncertainty or misinterpretation of single text items with bundle supervision and refinement (challenge 2).

The contribution of this paper can be summarized as follows. \ding{182} We introduce a new perspective that connects bundle structure and text-attributed graphs to provide robust supervision of graph neural networks. \ding{183} We propose a novel framework consisting of bundle sampling, bundle query, bundle supervision, and bundle refinement. We also provide rigorous theoretical analysis of our method, showing its tolerance to outlier nodes and the convergence properties of optimization. \ding{184} We perform extensive experiments on ten text-attributed graph datasets across various domains, and the results validate the effectiveness of the proposed method compared to competing baselines.

\section{Related Works}
\textit{\textbf{Text-Attributed Graphs.}} Text-attributed graphs (TAGs) are a special type of graphs whose nodes are associated with textual attributes \cite{yan2023comprehensive, zhang2024text}. They are common forms of data in many fields, such as citation networks \cite{berrebbi2022graphcite}, knowledge graphs \cite{peng2023knowledge}, social networks \cite{li2022distilling}, web page networks \cite{guo2021web}, \emph{etc.} Research on TAGs generally focuses on combining textual attributes with graph structures, with the help of text embedding methods \cite{wallach2006topic, meng2019spherical} and network embedding methods \cite{wang2016structural, wang2017community, liao2018attributed}. As the annotation costs of TAGs are usually high, efforts have been made in semi-supervised learning \cite{xiang2023semi, zhang2024taga}, transfer learning \cite{zhang2024collaborate, he2024unigraph, zhu2025graphclip}, and few-shot learning \cite{huang2023prompt, zhao2024pre, yu2025leveraging}. With the advancement of large language models, this work makes a step further, focusing on the zero-shot inference of text-attributed graphs \cite{chen2023label, wang2024llms, wang2025model} with the help of LLMs.

\textit{\textbf{Large Language Model for Graphs.}} Large language models (LLMs) \cite{yang2024qwen2, liu2024deepseek} have shown impressive performance in understanding data beyond natural languages, including programming languages \cite{jiang2024self}, sequences of numbers \cite{jiang2024empowering}, mathematics \cite{romera2024mathematical}, and graphs \cite{jin2024large, pan2024unifying, li2024glbench}. LLMs exhibit strong generalization ability, enabling few-shot or zero-shot inference on graphs. One line of research aims to build a foundation model, incorporating graph structures into current language model architectures \cite{xie2023graph, fatemi2023talk, lin2024langgfm,  wang2024llms}. These methods often require training to align the graph structure and natural language \cite{ zhang2024graphtranslator, zhu2025graphclip}, involving a large amount of labeled or paired data. Another line of research makes use of the inference capability of existing LLMs to generate labels or related information of graphs \cite{sun2023large, chen2023label, chen2024exploring, wang2025model}. However, they often use isolated nodes \cite{wang2025model} or explicit descriptions that are hard for LLMs to understand \cite{tang2024graphgpt}. Additionally, the noisy labels generated by LLMs can further harm subsequent inference operations (\emph{e.g.}, supervising neural networks, performing clustering) on graphs. Compared to these methods, this paper proposes to use text bundles to query LLMs and supervise graph neural networks, leading to richer information and more robust optimization.

\begin{figure}
    \centering
    \includegraphics[width=\linewidth]{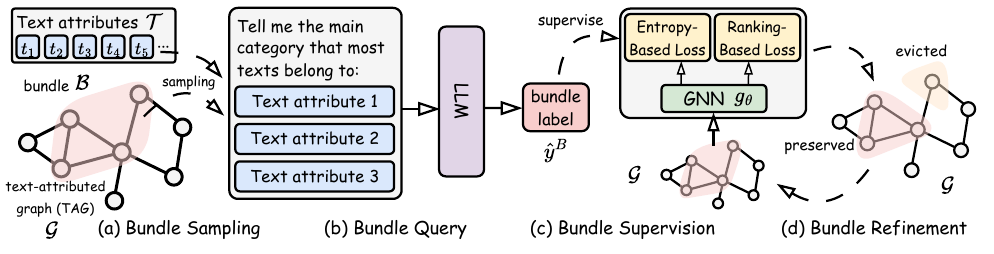}
    \vspace{-6mm}
    \caption{The overall framework of our method. We first sample nodes of proximity to form bundles (a), which are then used to query the LLM about their main categories (b). Subsequently, the bundle labels from the LLM's response are used to supervise a graph neural network (c). During optimization, we further refine the bundle to exclude noisy nodes (d).}
    \vspace{-4mm}
    \label{fig:framework}
\end{figure}

\section{Methodology}

\textit{\textbf{Problem Definition.}}
We denote a text-attributed graph as $\mathcal G=\langle \mathcal V, \mathcal E, \mathcal T, \mathcal Y\rangle$, where $\mathcal V$ is the set of nodes, $\mathcal E$ is the set of edges, $\mathcal T$ is the set of textual attributes, and $\mathcal Y$ is the set of node labels. Each node $v_i\in \mathcal V$ is associated with textual descriptions $t_i\in \mathcal T$ and the corresponding label $y_i$. For each node, we can obtain its vectorized embedding via a text encoder $f_\theta$, \emph{i.e.}, $\bm x_i = f_\theta (t_i)\in \mathbb R^d$. We denote the total number of nodes as $n=|\mathcal V|$. 
A node bundle is defined as a set of nodes in the graph, and a text bundle corresponding to the node bundle is defined as a set of text attributes associated with the node bundle. For simplicity, we use the term bundle and notation $\mathcal B$ to denote the indices of corresponding node bundles and text bundles.
The goal of zero-shot inference on text-attributed graphs is to infer the node labels $\mathcal Y$ according to the graph topology $\mathcal V$, $\mathcal E$, and the textual attributes $\mathcal T$.

\subsection{Framework Overview}
The overall framework of the proposed method is illustrated in Figure \ref{fig:framework}. We first perform bundle sampling, constructing node bundles according to topological or semantic proximity (Section \ref{sec:bundle_sampling}). With the obtained node bundles, we transform the corresponding text bundles into prompts and query the LLM about the most frequent category of the bundle (Section \ref{sec:bundle_query}). With these bundle labels, we perform bundle supervision, training graph neural networks with entropy-based and ranking-based supervision. Additionally, theoretical analysis is provided regarding the properties of bundle supervision to justify our design (Section \ref{sec:Bundle_Supervision}). During the optimization process, we further refine the bundles dynamically to exclude noisy components (Section \ref{sec:bundle_refinement}).

\subsection{Bundle Sampling}
\label{sec:bundle_sampling}
We first introduce the method for sampling bundles. Intuitively, we aim for most nodes within a bundle to belong to the same category (\emph{i.e.} a strong mode), so that LLMs more easily predict the mode category and the bundle label more accurately reflects the nodes it contains. To achieve this, we sample nodes of close proximity. Specifically, we first randomly sample the core node $v_c$ from the set of nodes $\mathcal V$, and then sample the rest of the nodes. We fix the size of a bundle as $n_B$, and design two criteria for sampling: topological proximity and semantic proximity.

\textit{\textbf{Topological Proximity.}} For a given core node $v_c$ in graph $\mathcal G$, a common assumption is that a node is similar to nodes topologically close to itself \cite{kipf2016semi, grover2016node2vec}. Formally, given two nodes $v_c$ and $v$, their topological proximity can be measured by the length of the shortest path from $v_c$ and $v$, denoted as $d^{\mathcal G}(v_c,v)$, and we can define topologically similar nodes with respect to $v_c$ as:
\begin{equation}\label{eq:topological-proximity}
\mathcal N^k_{\mathcal G}(v_c) = \left \{ i\mid 1 \le d^{\mathcal G}(v_i,v_c) \le k \right \},\quad k=\inf\left \{ x\mid |\mathcal N^x(v_c)|\ge n_B-1 \right \}
\end{equation}
where $k$ is an adaptive hop size. For core nodes with many ($k$-hop) neighbors, a smaller hop size is used, and vice versa. We then sample $(n_B-1)$ nodes from the neighborhood $\mathcal N^k_{\mathcal G}(v_c)$ to form the bundle $\mathcal B$ together with the original core node $v_c$.

\textit{\textbf{Semantic Proximity.}} For graphs with heterophily, topological proximity hardly entails similarity \cite{zhu2020beyond, zhu2021graph, zheng2024missing}. Therefore, we turn to semantic proximity utilizing vectorized representations of nodes. Specifically, given embeddings of each node $\mathcal X = \{\bm x_i\}_{i=1}^N$ and a core node $v_c$ with corresponding embedding $\bm x_c$, we construct the node bundle based on the closeness in the embedding space $\mathbb R^d$:
\begin{equation}\label{eq:semantic-proximity}
\mathcal B = \left\{ i \mid \bm x_i \in \mathcal N_{\mathcal X}^{n_B} (\bm x_c) \right\},
\end{equation}
where $\mathcal N_{\mathcal X}^{n_B} (\bm x_c)$ denotes the set of top $n_B$ vectors in $\mathcal X$ that are closest to $x_c$ in terms of Euclidean distance (\emph{i.e.}, $L_2$ distance) in the embedding space.

In practice, different criteria are adopted for different types of graphs. For graphs with high homophily (\emph{e.g.}, citation networks), topological proximity is used. For graphs with high heterophily (\emph{e.g.}, webpage networks), semantic proximity is adopted. We repeatedly sample a set of node bundles as $\{\mathcal B_1, \mathcal B_2, \dots, \mathcal B_{n_S}\}$, where $n_S$ is the number of bundles. For simplicity, we omit the subscript of bundles and use $\mathcal B$ for an arbitrary bundle in the following discussions.

\subsection{Bundle Query}
\label{sec:bundle_query}
We then query LLMs to obtain information about the bundles. While it might be straightforward to provide individual text attributes for node-level pseudo-labels, this approach carries the risk of limited information (as the LLMs only receive information from a single isolated node attribute) and unreliable responses (since the output pseudo-labels can be highly noisy). By using bundling, LLMs receive more information from proximate nodes, making the decision regarding the mode category easier than individual classification, which results in more reliable annotations.

With the node bundles selected, we obtain their corresponding text bundles and construct a single prompt $\mathcal P(\mathcal B)$ for each text bundle with dataset description and task description:
\begin{equation}\label{eq:prompt}
\mathcal P(\mathcal B)=\langle\textrm{dataset\_description}\rangle \operatorname{Concat}(\{t_i|i\in \mathcal B\}) \langle\textrm{task\_description}\rangle,
\end{equation}
where the $\textrm{Concat}(\cdot)$ operator concatenates all the text attributes in the bundle. We then query the LLM with the prompts to obtain the mode category of the bundle, denoted as $\hat{y}^B$.

\subsection{Bundle Supervision}\label{sec:Bundle_Supervision}
The bundle labels are then used to supervise a graph neural network. Since a bundle label represents the mode category that most nodes in the bundle belong to, nodes from other categories may also be included. Therefore, effective bundle supervision requires tolerance for these "outliers". To address this, we design two supervisions: entropy-based supervision and ranking-based supervision. We denote the graph neural network as $g_\theta$, and it generates probability distributions for each node as:
\begin{equation}
\{\bm z_i\}_{i=1}^n = g_\theta \left ( \{\bm x_i\}_{i=1}^n, \mathcal E \right ), \quad \bm p_i = \operatorname{softmax}(\bm z_i),
\end{equation}
where $\bm z_i\in\mathbb R^C$ is the logits, $\bm p_i\in\mathbb R^C$ is the probability, and $C$ is the number of classes.

\textit{\textbf{Entropy-based Supervision.}}
When a bundle $\mathcal B$ has label $\hat{y}^B$, the nodes in it are likely to fall into class $\hat{y}^B$ on average. Therefore, we compute the bundle class distribution $\bm p(\mathcal B)$ and the corresponding bundle-level entropy-based objective function $\mathcal L_{BE}$ as follows:
\begin{equation}\label{eq:bundle_ce}
\bm p(\mathcal B) = \operatorname{softmax}\left( \frac{1}{|\mathcal B|} \sum_{i\in \mathcal B} \bm z_i \right), \quad \mathcal L_{BE} = \operatorname{CE}\left(\bm p(\mathcal B), {\hat{y}^B}\right),
\end{equation}
where $\operatorname{CE}(\cdot,\cdot)$ is the cross-entropy loss. We then theoretically demonstrate that this bundle supervision (\emph{i.e.}, $\mathcal L_{BE}$) is more tolerant to outliers compared to individual supervision using cross-entropy. Formally, we have the following theorem:
\begin{theorem}\label{thm:tolerance}
Given a bundle $\mathcal B$, its corresponding bundle class distribution $\bm p(\mathcal B)=(p_1, p_2, \dots, p_C)$, an outlier node $v_o,o\in \mathcal B$ with probability distribution $\bm p_o=(p'_1, p'_2, \dots, p'_C)$, denote $m'=\operatorname{argmax}_i \{p'_i\}_{i=1}^C$. If the bundle label $\hat y\ne m'$, and $p'_{m'}\ge p_{m'}$, we have:
\begin{equation}
0 \le \frac{\partial \mathcal L_{BE}}{\partial \log p'_{m'}} \le \frac{\partial \mathcal L_{IE}}{\partial \log p'_{m'}}, ~~\textit{where}~~ \mathcal L_{BE} = \operatorname{CE}\left(\bm p(\mathcal B),{\hat{y}}\right) ~~\textit{and}~~ \mathcal L_{IE} = \frac{1}{|\mathcal B|}\cdot \operatorname{CE}\left(\bm p_o, {\hat{y}}\right),
\end{equation}
where $\hat y$ is the bundle label, $\mathcal L_{BE}$ is bundle supervision and $\mathcal L_{IE}$ is individual supervision.
\end{theorem}
\begin{remark}
Theorem \ref{thm:tolerance} suggests that when encountering "outlier" nodes that conflict with the predicted mode category and the bundle distribution (i.e., the condition $\hat y\ne m'$ and $p'_{m'}\ge p_{m'}$ in the theorem), the bundle cross-entropy objective function (i.e., $\mathcal L_{BE}$ defined in Eq. \ref{eq:bundle_ce}) is more tolerant compared to supervising the nodes in the bundle individually (\emph{i.e.}, $\mathcal L_{IE}$ defined in the theorem), as evidenced by a smaller penalty imposed by the gradient.
\end{remark}

\textit{\textbf{Ranking-based Supervision.}} To ensure that the supervision focuses more on bundles where the predicted bundle labels do not dominate the bundle's bundle probability distribution, we adopt the concept of ranking loss \cite{chen2009ranking, menon2012predicting, wang2019ranked}, and design a ranking-based loss as follows:
\begin{equation}\label{eq:ranking-loss}
\mathcal L_R = -\min\left(\log \bm p(\mathcal B)_{\hat{y}^B} - \log \max_{i=1}^C \left\{ \bm p(\mathcal B)_i \right\}, 0\right),
\end{equation}
where $p(\mathcal B)_i\in \mathbb R$ denotes the $i$-th component of vector $p(\mathcal B)$ (\emph{i.e.}, the predicted probability of class $i$). When the category of the bundle label $\hat{y}^B$ is not the highest in the predicted bundle probability distribution by the GNN $g_\theta$, the bundle $\mathcal B$ is penalized by this loss function. On the other hand, when the category of $\hat{y}^B$ has a high bundle probability $\bm p(\mathcal B)_{\hat{y}^B}$ (which does not necessitate all $\bm z_i,i\in\mathcal B$ to be high), the loss will be zero. In our implementation, a combination of the two supervision objectives is used, leading to the final objective function as follows:
\begin{equation}\label{eq:final-loss}
\mathcal L = \mathcal L_{BE} + \mathcal L_R.
\end{equation}

\textit{\textbf{Theoretical Analysis.}}

We then aim to present a rigorous theoretical analysis of the proposed method, focusing in particular on the convergence of the bundle supervision process.
Before going into the details, we first examine the smoothness of our entropy-based objective $\mathcal{L}_{BE}$. More specifically, we have the following results:
\begin{theorem}\label{thm2}Given a graph neural network $g_{\theta}$, if its corresponding first-order and second-order partial derivatives are bounded, that is, $\|\nabla z_{i,c}(\theta)\|_{\infty}\le G$ and $\max(\left|\nabla^{2}z_{i,c}(\theta)\right|)\le M$ where $z_{i,c}$ is the `$c$'-th logit of the output vector $\bm{z}_{i}\triangleq(z_{i,1},\dots,z_{i,C})$ provided by GNN $g_{\theta}$, then we can show that the cross-entropy loss function $\mathcal{L}_{BE}(\theta)$ defined in Eq. \ref{eq:bundle_ce} satisfies the following conditions:
\begin{enumerate}
    \item[\textbf{i):}] The cross-entropy loss function $\mathcal{L}_{BE}(\theta)$ has a bounded gradient, i.e., $\|\nabla\mathcal{L}_{BE}(\theta)\|_{\infty}\le\frac{2G}{|\mathcal{B}|}$ where the symbol $|\mathcal{B}|$ represents the cardinality of bundle $\mathcal{B}$; 
    \item[\textbf{ii):}] The second-order partial derivatives of the cross-entropy loss function $\mathcal{L}_{BE}(\theta)$ is also bounded, namely, $\max\left(\left|\nabla^{2}\mathcal{L}_{BE}(\theta)\right|\right)\le\frac{2(M+G^{2})}{|\mathcal{B}|}$, which simultaneously means the loss $\mathcal{L}_{BE}(\theta)$ is $\left(\frac{2n_d(M+G^{2})}{|\mathcal{B}|}\right)$-smooth, that is, 
    \begin{equation*}
        \|\nabla\mathcal{L}_{BE}(\theta_{1})-\nabla\mathcal{L}_{BE}(\theta_{2})\|_{2}\le\frac{2n_d(M+G^{2})}{|\mathcal{B}|}\|\theta_{1}-\theta_{2}\|_{2},
    \end{equation*} where $n_d$ is the dimension of the unknown parameter $\theta$.
\end{enumerate}
\end{theorem}
\begin{remark}
It is worth noting that, in Theorem~\ref{thm2}, the symbol $\max\left(|\bf M|\right)$ represents the maximum absolute value among the elements of matrix $\bf M$. Moreover, $\|\cdot\|_{2}$ and $\|\cdot\|_{\infty}$ denote the standard $L_{2}$ norm and $L_{\infty}$ norm, respectively.
\end{remark}
\begin{remark}
Theorem~\ref{thm2} indicates that the smoothness and differentiability of the graph neural network $g_{\theta}$ can, to some extent, be inherited by our adopted cross-entropy loss function $\mathcal{L}_{BE}$. 
\end{remark}

With the results of Theorem~\ref{thm2}, we next show that, under some mild conditions, the commonly used gradient descent algorithm for training GNN $g_{\theta}$ can finally converge to a stationary point of our adopted loss function $\mathcal L\triangleq(\mathcal L_{BE} + \mathcal L_R)$. Before that, we first characterize the dynamics of the general gradient descent algorithm, namely, we suppose $\theta_{t+1}\triangleq\theta_{t}-\eta\nabla\mathcal{L}_{R}$ where $\eta>0$ is the learning rate and the time $t\in\{1,2,\dots,T\}$. Subsequently, we present the detailed results regarding the convergence of our adopted bundle supervision process, that is,
\begin{theorem}\label{thm3}   
Under the assumptions of Theorem~\ref{thm2} and the condition $\eta<\frac{|\mathcal{B}|}{n_d(M+G^{2})}$, if, when the iteration index $t$ is large, the model parameter $\theta_{t}$ provided by gradient descent algorithm can effectively fit the predicted bundle label $\hat{y}^B$, namely, $\hat{y}^B\in\mathop{\arg\max_{i\in\{1,\dots,C\}}}\left\{ \bm p_{\theta_{t}}(\mathcal B)_i \right\}$, then we can verify that the final obtained model parameter $\theta_{T+1}$ will converge to a stationary point of the adopted loss function $\mathcal{L}(\theta)$, that is to say, $\|\nabla\mathcal{L}(\theta_{T+1})\|_{2}$ can approach toward a small value as $T\rightarrow\infty$.
\end{theorem}
\begin{remark}
It is important to emphasize that when our GNN model $g_{\theta}$ possesses certain structural properties, extensive research has shown that the resulting stationary point of the aforementioned gradient descent algorithm can exhibit strong generalization capabilities~\citep{krishnagopal2023graph,bos2022convergence} and in some cases, may even correspond to a global minimum~\citep{rychlik2019proof,mousavi-hosseini2025learning,mei2018mean}.
\end{remark}

\subsection{Bundle Refinement}
\label{sec:bundle_refinement}
During optimization of the graph neural network $g_\theta$, the bundle $\mathcal B$ may include nodes that do not belong to the category of $\hat{y}^B$. To address this, we design the bundle refinement process that excludes these noisy nodes by evicting those with lower confidence in class $\hat{y}^B$. Specifically, given the node-level probability distribution in a bundle, \emph{i.e.}, $\bm p_i,i\in \mathcal B$, we denote the confidence of $\bm p_i$ with respect to class $\hat{y}^B$ as $\bm p_{i,\hat{y}^B}$. We evict the less confident node in the bundle as:
\begin{equation}\label{eq:refinement}
\mathcal B \leftarrow \left\{ i\mid i\in\mathcal B ~\land~ \bm p_{i,\hat{y}^B} > \min_{j\in \mathcal B} \bm p_{j,\hat{y}^B} \right\},
\end{equation}
where $\leftarrow$ denotes the update of the bundle. Bundle refinement is performed multiple times during the optimization process of $g_\theta$. By evicting the less confident nodes that are potentially misaligned with the bundle label, the noise in bundle supervision is further reduced. Through bundle refinement, the initial bundles, sampled via topological or semantic proximity, are dynamically adjusted during the supervision of the graph neural network to fulfill the predicted bundle label $\hat y^B$ queried from the LLM, making the proposed method robust. 

\section{Experiments}
\subsection{Experimental Setup}\label{sec:setup}
\textit{\textbf{Datasets.}} In the experiments, we use ten representative datasets, \emph{i.e.}, Cora \cite{mccallum2000automating}, CiteSeer \cite{giles1998citeseer}, Wikics \cite{mernyei2020wiki}, History \cite{ni2019justifying}, Children \cite{ni2019justifying}, Sportsfit \cite{ni2019justifying}, Cornell \cite{craven1998learning}, Texas \cite{craven1998learning}, Wisconsin \cite{craven1998learning}, and Washington \cite{craven1998learning}. Among these datasets, Cora and CiteSeer are citation networks. Wikics is a knowledge graph derived from Wikipedia. History, Children, and Sportsfit are e-commerce networks of different types of products (\emph{i.e.}, history books, children's literature, sports goods). Cornell, Texas, Wisconsin, and Washington are web page networks of universities. The datasets cover both homophilic and heterophilic graphs, with the first six datasets of high homophily and the last four of low homophily.

\textit{\textbf{Compared Baselines.}} We compare a spectrum of methods with our method. The compared methods include the following categories. \emph{$\blacktriangleright$ Text encoders}, including SBERT \cite{reimers2019sentence}, RoBERTa \cite{liu2019roberta}, OpenAI's Text-Embedding-3-Large (TE-3-Large) \cite{openai2024textembedding} and LLM2Vec \cite{behnamghader2024llm2vec}. \emph{$\blacktriangleright$ Generative LLMs}, including GPT-3.5-turbo \cite{achiam2023gpt} and GPT-4o \cite{hurst2024gpt}. \emph{$\blacktriangleright$ Graph self-supervised learning methods}, including DGI \cite{velivckovic2018deep} and GraphMAE \cite{hou2022graphmae}. \emph{$\blacktriangleright$ Graph foundation models or graph learning methods with LLMs}, including OFA \cite{liu2023one}, GOFA \cite{kong2024gofa}, UniGLM \cite{fang2024uniglm}, ZeroG \cite{li2024zerog}, GraphGPT \cite{tang2024graphgpt}, LLAGA \cite{chen2024llaga}, and LLM-BP \cite{wang2025model}.

\textit{\textbf{Implementation Details.}} In the experiments, we use GPT-4o \cite{hurst2024gpt} as the default LLM for bundle query. In bundle sampling, we set the bundle size $n_B$ as 5 and the number of bundles $n_S$ as 100 for all datasets. We train the GNN on an NVIDIA RTX 3090 GPU.

\begin{table}[h]
    \centering
    \caption{Prediction accuracies of our method compared to baselines across datasets. We mark the best results in \textbf{bold} and the second-best with \underline{underline}.}
    \vspace{-1mm}
    \resizebox{\textwidth}{!}{%
    \def\arraystretch{1.1}
    \begin{tabular}{l *{10}{>{\centering\arraybackslash}p{1.1cm}}}
\Xhline{1.2pt}
\rowcolor{tableheadcolor!20}
Method & Cora & CiteSeer & WikiCS & History & Children & Sportsfit & Cornell & Texas & Wisc. & Wash. \\
\Xhline{1.0pt}
SBERT & 69.75 & 66.69 & 59.06 & 53.53 & 22.59 & 43.79 & 63.66 & 64.58 & 62.10 & 63.52 \\
\rowcolor{tablerowcolor} RoBERTa & 70.71 & 66.95 & 59.08 & 55.39 & 24.25 & 41.51 & 61.68 & 62.25 & 60.33 & 60.60 \\
TE-3-Large & 71.90 & 66.24 & 61.78 & 50.15 & 24.68 & 58.39 & 81.50 & 75.42 & 73.14 & 66.35 \\
\rowcolor{tablerowcolor} LLM2Vec & 67.34 & 67.13 & 62.34 & 53.14 & 25.56 & 57.00 & 81.26 & 76.68 & 73.36 & 65.92 \\
GPT-3.5-turbo & 70.11 & 66.83 & 65.53 & 55.07 & 29.73 & \underline{67.21} & 45.54 & 56.14 & 58.86 & 51.09 \\
\rowcolor{tablerowcolor} GPT-4o & 70.29 & 64.77 & 66.10 & 53.30 & \underline{30.76} & 66.35 & 45.54 & 63.10 & 56.60 & 48.90 \\
DGI & 16.79 & 15.24 & 14.98 & 20.98 & 2.22 & 7.48 & 14.66 & 11.23 & 12.08 & 20.96 \\
\rowcolor{tablerowcolor} GraphMAE & 15.13 & 8.11 & 8.91 & 36.36 & 7.24 & 30.50 & 23.04 & 17.65 & 23.02 & 24.89 \\
OFA & 20.36 & 41.31 & 30.77 & 8.25 & 3.05 & 15.18 & 29.84 & 11.77 & 4.80 & 6.04 \\
\rowcolor{tablerowcolor} GOFA & 71.06 & 65.72 & \underline{68.62} & 56.25 & 12.15 & 37.87 & 39.50 & 38.37 & 32.51 & 31.02 \\
UniGLM & 45.57 & 52.26 & 55.05 & 44.24 & 21.48 & 33.46 & 23.03 & 21.39 & 27.16 & 24.01 \\
\rowcolor{tablerowcolor} ZeroG & 60.40 & 50.35 & 46.74 & 36.55 & 12.72 & 14.27 & 10.47 & 53.48 & 12.66 & 8.30 \\
GraphGPT & 17.48 & 13.93 & 33.59 & 12.31 & 9.94 & 4.53 & 10.18 & 18.48 & 12.35 & 20.64 \\
\rowcolor{tablerowcolor} LLAGA & 11.62 & 19.52 & 10.98 & 7.95 & 10.09 & 1.84 & 12.57 & 15.51 & 15.09 & 10.48 \\
LLM-BP & \underline{72.59} & \underline{69.51} & 67.75 & \underline{59.86} & 24.81 & 61.92 & \underline{83.28} & \underline{81.66} & \underline{77.75} & \underline{73.14} \\
\Xhline{1.0pt}
\rowcolor{ourrowname} \method{} (ours) & \textbf{75.09} & \textbf{72.37} & \textbf{71.03} & \textbf{67.31} & \textbf{31.75} & \textbf{75.88} & \textbf{84.82} & \textbf{92.51} & \textbf{87.17} & \textbf{81.66} \\
\Xhline{1.2pt}
    \end{tabular}
    }%
    \vspace{-2mm}
    \label{tab:mainresults}
\end{table}
\subsection{Main Results}
\textbf{\textit{Comparison with Existing Methods.}}
We compare our method against 15 baselines across 10 datasets in Table \ref{tab:mainresults}. From the results, we can see that our method consistently outperforms competitive baselines in all 10 datasets, showing the effectiveness of the proposed text bundling method. Text embedding methods (\emph{e.g.}, SBERT, LLM2Vec) and generative LLMs (\emph{e.g.}, GPT-4o) achieve moderate performance on many datasets. However, their ignorance of the graph topology leads to weaker performance, especially when the structures are important. Graph self-supervised learning methods (\emph{e.g.}, DGI, GraphMAE) generally yield low accuracy without the assistance of LLMs and their strong generalization capability. For foundation models (\emph{e.g.}, GOFA, ZeroG) that incorporate graphs in LLMs for joint training, their high performance is not consistent, worsening with graphs out of their original training distribution (\emph{e.g.}, in university web page networks). By comparison, our method consistently outperforms baselines on various datasets covering different domains. Additionally, our method is agnostic to the specific architecture of the graph neural network, allowing us to flexibly benefit from the advancement of GNN architectures when facing different types of graph structures (\emph{e.g.}, homophilic graphs and heterophilic graphs).

\begin{figure}[ht]
\vspace{-2mm}
  \begin{minipage}{0.49\textwidth}
  \centering
  \begin{table}[H]
    \centering
    \caption{The prediction accuracies under different LLM backbones on four datasets. The best is marked in \textbf{bold} and the second-best \underline{underline}.}
    \vspace{-1mm}
    \resizebox{\textwidth}{!}{%
    \def\arraystretch{1.15}
    \begin{tabular}{l *{4}{>{\centering\arraybackslash}p{1.1cm}}}
\Xhline{1.2pt}
\rowcolor{tableheadcolor!20}
LLM & Cora & History & Sportsfit & Texas \\
\Xhline{1.0pt}
GPT-4o & \underline{75.09} & 67.31 & \textbf{75.88} & \underline{92.51} \\
\rowcolor{tablerowcolor} GPT-3.5-turbo & 73.25 & 69.87 & 69.82 & 89.30 \\
GPT-4.1-nano & 70.11 & \textbf{71.09} & 66.11 & 90.37 \\
\rowcolor{tablerowcolor} Deepseek-V3 & \textbf{75.28} & 67.00 & 73.52 & 85.56 \\
Gemini-2.5-flash & 73.25 & \underline{70.08} & \underline{74.98} & \textbf{93.05} \\
\Xhline{1.2pt}
    \end{tabular}
    }%
    \label{tab:backbones}
\end{table}
  \end{minipage}
  \hfill
  \begin{minipage}{0.49\textwidth}
    \centering
    \begin{table}[H]
    \centering
    \caption{Ablation studies on four datasets.}
    \vspace{-1mm}
    \resizebox{\textwidth}{!}{%
    \def\arraystretch{1.15}
    \begin{tabular}{l *{4}{>{\centering\arraybackslash}p{1.1cm}}}
\Xhline{1.2pt}
\rowcolor{tableheadcolor!20}
Method & Cora & History & Sportsfit & Texas \\
\Xhline{1.0pt}
V1: R.S. & 70.48 & 61.80 & 65.60 & 88.24 \\
\rowcolor{tablerowcolor} V2: I.Q. & 71.96 & 63.95 & 72.61 & 84.49 \\
V3: \emph{w/o} $\mathcal L_{BE}$ & 70.11 & 64.49 & 65.29 & 91.44 \\
\rowcolor{tablerowcolor} V4: \emph{w/o} $\mathcal L_R$ & 73.99 & 66.73 & 75.48 & 86.10 \\
V5: \emph{w/} $\mathcal L_{IE}$ & 73.43 & 66.29 & 74.05 & 85.03 \\ 
\rowcolor{tablerowcolor} V6: \emph{w/o} B.R. & 73.89 & 66.55 & 73.00 & 91.98 \\
\Xhline{1.0pt}
\rowcolor{ourrowname} \method{} (ours) & \textbf{75.09} & \textbf{67.31} & \textbf{75.88} & \textbf{92.51} \\
\Xhline{1.2pt}
    \end{tabular}
    }%
    \label{tab:ablation}
\end{table}
  \end{minipage}
  \vspace{2mm}
\end{figure}

\textit{\textbf{Performance Under Different LLM Backbones.}}
We also show the prediction accuracies of our method using different LLMs. Specifically, we provide results on five LLMs, including GPT-4o \cite{hurst2024gpt} (used as default), GPT-3.5-turbo \cite{achiam2023gpt}, GPT-4.1-nano \cite{openai2024gpt41nano}, Deepseek-V3 \cite{liu2024deepseek}, and Gemini-2.5-flash \cite{google2024gemini25flash}. The results on four datasets (\emph{i.e.}, Cora, History, Sportsfit, Texas) are shown in Table \ref{tab:backbones}. As can be seen from the results, using alternative LLMs generally yields satisfactory performance on average. Among these LLMs, GPT-4o and Gemini-2.5-flash perform relatively better, while cheaper or older LLMs like GPT-3.5-turbo, GPT-4.1-nano, Deepseek-V3 yield decent accuracies as well. This suggests that our method can benefit from the advancement of LLMs.

\subsection{Ablation Studies}
We investigate how the different mechanisms used in our method affect the final accuracy, and we present the ablation studies in Table \ref{tab:ablation}. We construct a set of variants of our method (marked as V1 to V6): V1 uses random sampling (R.S.) instead of topological proximity or semantic proximity to obtain bundles. V2 uses individual query (I.Q.), asking the LLM about the category of each node with the text attribute. V3 removes the entropy-based loss $\mathcal L_{BE}$. V4 removes the ranking-based loss $\mathcal L_R$. V5 uses individual supervision, \emph{i.e.}, $\mathcal L_{IE}$ defined in Theroem \ref{thm:tolerance}. V6 does not employ bundle refinement. As can be seen from the results, each technique proposed is helpful for the overall accuracy, and removing them causes performance degradation. Additionally, we find that bundle sampling is important, especially when the number of classes is large (in this case, 12 classes for History and 13 classes for Sportsfit, both of which witness a severe drop in accuracy with random bundle sampling). One explanation is that inappropriate sampling causes the nodes in a bundle to be more uniformly distributed across various categories, making it difficult to decide the bundle class (with a weaker mode category) and perform bundle supervision (with noisier bundle labels). Moreover, we find that individual supervision ($\mathcal L_{IE}$) is weaker than bundle supervision, which suggests that our supervision method is more tolerant to bundle outliers.

\begin{figure}
    \centering
    \includegraphics[width=\linewidth]{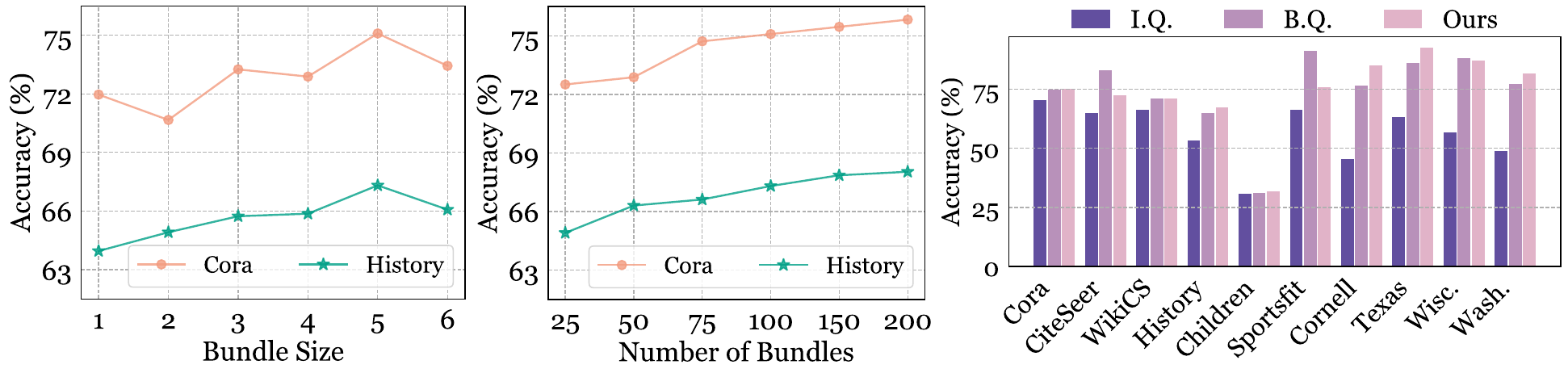}
    \vspace{-4mm}
    \caption{\textbf{Left}: prediction accuracies under different bundle sizes (\emph{i.e.}, $n_B$). \textbf{Middle}: prediction accuracies with different numbers of bundles (\emph{i.e.}, $n_S$). \textbf{Right}: accuracy comparison of individual query (I.Q.), bundle query (B.Q.), and our method (Ours).}
    \vspace{-2mm}
    \label{fig:sensitivity-comparison}
\end{figure}

\subsection{Hyperparameter Analysis}
\textit{\textbf{Effects of Bundle Size $n_B$.}} We also show the prediction accuracies using different bundle sizes $n_B$ in Figure \ref{fig:sensitivity-comparison} (Left). As can be seen from the figure, setting $n_B$ to 5 achieves relatively good performance. Smaller bundle sizes make it difficult to provide sufficient information about the neighborhood, while larger bundle sizes have the risk of including categories other than the mode category. Moreover, we observe that odd bundle sizes are relatively better than even ones. A possible explanation is that even bundles are more likely to have ties (\emph{e.g.}, two nodes from class A and two from class B in a four-node bundle), causing confusion and noisy supervision. This is less significant in datasets with a larger number of classes (\emph{e.g.}, the History dataset with 12 classes), where odd bundles may also include a node from the third class, failing to resolve the ties.

\textit{\textbf{Effects of the Number of Bundles $n_S$.}} We also show the prediction accuracies with different numbers of bundles in Figure \ref{fig:sensitivity-comparison} (Middle). As can be seen from the figure, larger numbers of bundles generally yield better accuracy. As one bundle corresponds to one query of the LLM, increasing the number of bundles indicates additional feedback from the LLM, leading to additional supervision signals that better cover the graph topology and bring robustness from redundancy. Nevertheless, additional queries incur additional computation resources. Balancing performance and cost from querying LLMs, we set the number of bundles $n_S$ to 100 in the experiments.

\begin{figure}
    \centering
    \includegraphics[width=\linewidth]{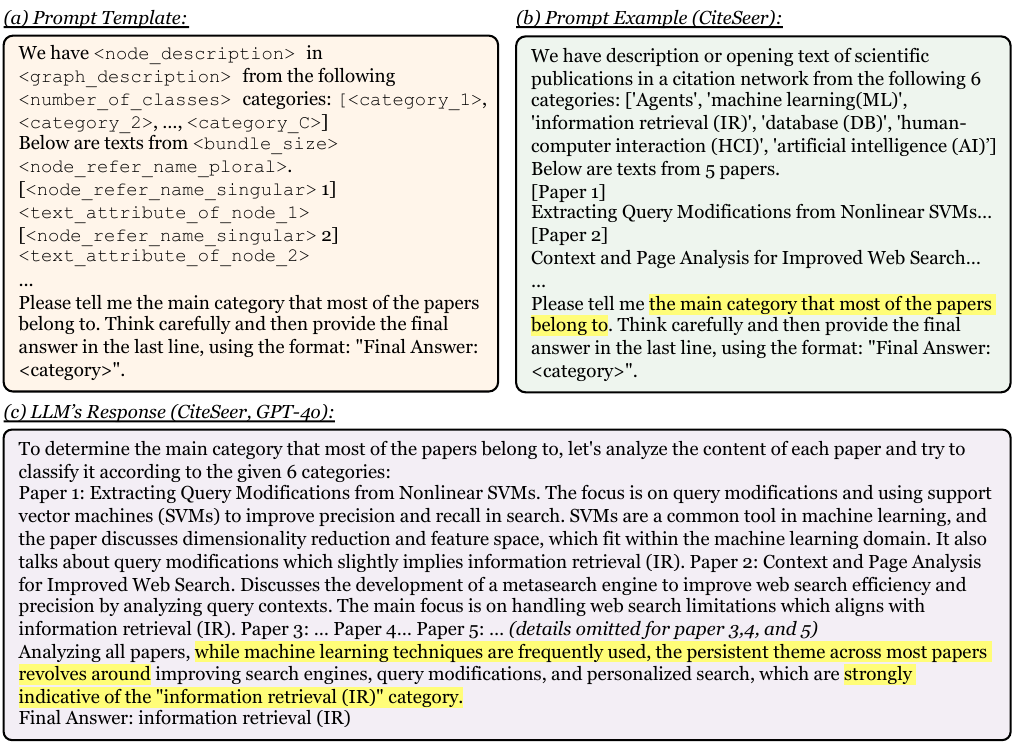}
    \vspace{-5mm}
    \caption{The prompt template of bundle query (a), an example of the prompt on the CiteSeer dataset (b), and an example of the response of GPT-4o to the query (c).}
    \vspace{-5mm}
    \label{fig:prompt}
\end{figure}

\subsection{Further Analysis}
\textit{\textbf{Bundle Query v.s. Individual Query.}}
We then show that the LLM can better handle bundle queries than individual queries. Specifically, we provide the LLM's classification accuracies given individual queries (marked as I.Q.) and bundle queries (marked as B.Q.) in Figure \ref{fig:sensitivity-comparison} (Right). From the figure, we can see that predicting the main category of the text bundles is generally easier than classifying individual text items, and in some datasets (\emph{e.g.}, CiteSeer, Cornell), the improvement is fairly large. We also show the overall prediction accuracies of our method, and we can see a general connection between the improvement of bundle queries and our method (compared to individual queries). This shows that the proposed text bundling method increases the reliability and robustness of supervision signals from LLMs and thereby improves the overall performance.

\textit{\textbf{Prompt Examples and the LLM's Response.}} We also provide the prompt template, an example of the prompt, and the LLM's response in Figure \ref{fig:prompt}. In the prompt, we provide information about the nodes and the graphs. We then ask the LLM to find the main category that most of the papers in the text bundle belong to. For the LLM's response, we can see that although machine learning is a frequent topic of research among the papers in the bundle, the LLM discovers a "\emph{persistent theme across most papers}" to be strongly related to information retrieval. Without text bundling, the LLM may hesitate between machine learning and information retrieval when classifying Paper 1, as its analysis suggests that this paper "fits within the machine learning domain" and also "slightly implies information retrieval". Such ambiguity would cause noise in classification results and be harmful for potential subsequent operations (\emph{e.g.}, clustering, supervision of GNNs). By comparison, our method allows the LLM to obtain more information, finding a persistent theme that represents most text items in the bundle, improving the reliability of LLM's response.

\section{Conclusion}
This paper investigates the important problem of zero-shot inference on text-attributed graphs with the help of LLMs. While previous efforts suffer from limited information on graph structure and unreliable responses, this paper proposes a novel method named \methodlower{} that queries the LLM with text bundles to obtain bundle-level labels. Subsequently, the bundle labels are used to supervise a graph neural network, which is then used for classification. We provide theoretical analysis of our method, showing its tolerance of outlier nodes in the bundle and the convergence properties of optimization. We further refine the nodes in the bundle to exclude noisy items. Extensive experiments are performed on ten datasets across different domains against a number of competing baselines, and the results confirm the effectiveness of the proposed method.

\textbf{Limitations and Broader Impacts.}
This paper focuses on text-attributed graphs, where each node is associated with a textual attribute. For graphs where node attributes are hard for LLMs to understand, the proposed text bundling method is not directly applicable. For graph structures on which GNNs are inherently weak or inferior to alternatives, this method may not be directly applicable.
As for broader impacts, the proposed text bundling method improves the zero-shot inference ability of LLMs on text-attributed graphs, facilitating downstream applications in many fields, including social network analysis, recommendation systems, web page analysis, and knowledge graph understanding.


\bibliographystyle{plain}
\bibliography{main.bib}

\end{document}